\documentclass[letterpaper]{article} % DO NOT CHANGE THIS
\usepackage{aaai23}  % DO NOT CHANGE THIS
\usepackage{times}  % DO NOT CHANGE THIS
\usepackage{helvet}  % DO NOT CHANGE THIS
\usepackage{courier}  % DO NOT CHANGE THIS
\usepackage[hyphens]{url}  % DO NOT CHANGE THIS
\usepackage{graphicx} % DO NOT CHANGE THIS
\usepackage{natbib}  % DO NOT CHANGE THIS AND DO NOT ADD ANY OPTIONS TO IT
\usepackage{caption} % DO NOT CHANGE THIS AND DO NOT ADD ANY OPTIONS TO IT
\usepackage{booktabs}
\usepackage{bm}
\usepackage[usenames,dvipsnames]{color} 
\usepackage{subfigure}

\usepackage{algorithm}
\usepackage[noend]{algorithmic}
\usepackage{multirow}

\usepackage{amsfonts}

\usepackage{amsmath}
\usepackage{pifont}

\usepackage{newfloat}
\usepackage{listings}
\DeclareCaptionStyle{ruled}{labelfont=normalfont,labelsep=colon,strut=off} % DO NOT CHANGE THIS
\floatstyle{ruled}
\newfloat{listing}{tb}{lst}{}
\floatname{listing}{Listing}
\title{GOHSP: A Unified Framework of Graph and Optimization-based Heterogeneous Structured Pruning for Vision Transformer}
\author {
    Miao Yin\textsuperscript{\rm 1}\footnote{This work was done during Miao Yin's internship at Samsung Research America.}
    Burak Uzkent\textsuperscript{\rm 2},
    Yilin Shen\textsuperscript{\rm 2},
    Hongxia Jin\textsuperscript{\rm 2},
    Bo Yuan\textsuperscript{\rm 1}
}
\affiliations {
    \textsuperscript{\rm 1} Rutgers University, 
    \textsuperscript{\rm 2} Samsung Research America\\
}
\usepackage{bibentry}
\begin{document}

\maketitle

\begin{abstract}
The recently proposed Vision transformers (ViTs) have shown very impressive empirical performance in various computer vision tasks, and they are viewed as an important type of foundation model. However, ViTs are typically constructed with large-scale sizes, which then severely hinder their potential deployment in many practical resources-constrained applications. To mitigate this challenging problem, structured pruning is a promising solution to compress model size and enable practical efficiency. However, unlike its current popularity for CNNs and RNNs, structured pruning for ViT models is little explored.

In this paper, we propose GOHSP, a unified framework of Graph and Optimization-based Structured Pruning for ViT models. We first develop a graph-based ranking for measuring the importance of attention heads, and the extracted importance information is further integrated to an optimization-based procedure to impose the heterogeneous structured sparsity patterns on the ViT models. Experimental results show that our proposed GOHSP demonstrates excellent compression performance. On CIFAR-10 dataset, our approach can bring $40\%$ parameters reduction with no accuracy loss for ViT-Small model. On ImageNet dataset, with $30\%$ and $35\%$ sparsity ratio for DeiT-Tiny and DeiT-Small models, our approach achieves $1.65\%$ and $0.76\%$ accuracy increase over the existing structured pruning methods, respectively.

% test GOHSP on ImageNet and show that it compresses DeiT-Tiny (5M) and DeiT-Small (22M) models by $30\%$ and $40\%$ while suffering minimal accuracy loss. Additionally, GOHSP achieves $40\%$ compression on ViT-base model on CIFAR10 without loss in accuracy.XXXXXX
% Finally, we fine-tune the pruned model to minimize the loss in accuracy compared to the original model.
% Additionally, we propose self-distillation method to better train the compressed network. 

%XXXXXX
\end{abstract}

\section{Introduction}

Recently applying transformer architecture to computer vision has emerged as an important forefront of foundation model design~\cite{dosovitskiy2020image}. Thanks to the delicate vision-specific self-attention, inherent minimal inductive biases and high scalability and parallelism, \textit{vision transformers (ViTs)}~\cite{dosovitskiy2020image,touvron2021training,zhou2021deepvit} have shown very outstanding and even state-of-the-art performance in many fundamental and downstream image and video processing tasks, such as image classification, object detection, super-resolution, video classification etc.

Motivated by the scaling success of the giant natural language processing (NLP) transformers (e.g., BERT~\cite{devlin2018bert} and GPT-3~\cite{brown2020language}), the existing ViTs are also constructed with large model sizes to adapt for massive data training~\cite{zhai2021scaling}. Consequently, they are suffering from huge memory footprints and extensive computational costs. These limitations, if not being properly addressed, could severely hinder the widespread adoption of ViTs in many practical scenarios, especially on the resource-constrained mobile platforms and Internet-of-things (IoT) devices.

To mitigate this challenging problem, one attractive solution is to perform model compression~\cite{yu2017compressing,kim2015compression,pan2019compressing} to reduce the network costs without affecting task performance. However, unlike the current popularity of compressing convolutional and recurrent neural networks (CNNs and RNNs), ViT-oriented model compression has not been systematically studied yet. In particular, \textit{structured pruning}, as an important hardware-friendly compression strategy that can bring practical efficiency on the off-the-shelf hardware, is little explored for ViT models.

To date, a rich set of structured pruning approaches have been proposed and investigated in the existing literatures, and most of them focus on sparsifying the CNNs at the channel level~\cite{he2017channel,ye2018rethinking}. On the other hand, as will be analyzed and elaborated in Section \ref{sec:analysis}, because of the difference of the underlying architecture, the structured sparse ViT models can exhibit multi-granularity sparsity (i.e., head-level and column-level) in the different component modules (i.e., attention head and multi-layer perception (MLP)). The co-existence of such heterogeneous sparse patterns raises a series of new research challenges and questions when we consider the efficient structured pruning strategy for ViT models. For instance, for each component module what is the corresponding suitable pruning criterion to obtain its specific sparse pattern? Also, how should we perform the entire pruning process across different modules with different levels of granularity sparsity to optimize the overall compression and task performance?

\textbf{Technical Preview \& Contributions.} To answer these questions, in this paper we propose GOHSP, a unified framework of Graph and Optimization-based Structure Pruning for vision transformer. To be specific, we first develop a graph-based ranking approach to measure the importance of attention heads. As a soft-pruning guideline, such importance information is then integrated to the overall optimization-based procedure to impose the different types of structured sparsity in a joint and global way. Overall, the contributions of this paper are summarized as follows:

\begin{itemize} 
    \item We propose a graph-based ranking algorithm to measure and determine the importance of attention heads. By modeling the inter-head correlation as a converged Markov chain, the head importance can be interpreted and calculated as the stationary distribution, which is further used as a soft guideline for the overall pruning procedure.
    \item We propose a unified framework to jointly optimize different types of structured sparsity in the different modules. The complicated coordination for different sparse patterns are automatically learned and optimized in a systematic way.
    \item We evaluate the performance of our structured pruning approach of different ViT models on different datasets. On CIFAR-10 dataset, our approach can bring $40\%$ parameters reduction with no accuracy loss for ViT-Small model. On ImageNet dataset, with $30\%$ and $40\%$ sparsity ratio for DeiT-Tiny and DeiT-Small models, our approach achieves $1.65\%$ and $0.76\%$ accuracy increase than the existing structured pruning methods, respectively.
\end{itemize}

\section{Related Work}
\textbf{Vision Transformer.} Inspired by the grand success of transformer architecture in NLP domains, deep learning researchers have actively explored the efficient transformer-based neural networks for computer vision. Most recently, several vision transformers (ViTs) and their variants have already shown very impressive performance in several image and video processing tasks~\cite{dosovitskiy2020image,touvron2021training,zhou2021deepvit}. However, in order to achieve competitive performance with the state-of-the-art CNNs, ViTs typically have to scale up their model sizes and therefore they suffer from costly computation and storage.

\textbf{Dynamic Inference with ViTs.} To reduce the deployment costs of ViTs, several works~\cite{wang2021not,bakhtiarnia2021single,rao2021dynamicvit,meng2022adavit,xu2022evo,uzkent2020efficient,uzkent2020learning} have been proposed to improve the processing speed via dynamically pruning the tokens/patches or skipping transformer components adaptively. Essentially as dynamic inference approaches, this set of works do not pursue to reduce the model sizes but focus on input-aware inference to obtain practical speedup. Our structured pruning-based solution is orthogonal to them, and these two different strategies can be potentially combined together to achieve higher speed and smaller memory footprint.

% Vision transformer (ViT) \cite{dosovitskiy2020image} adopts attention-based transformer \cite{vaswani2017attention} to process visual data and shows competitive performance with state-of-the-art CNNs when it is pre-trained on large scale dataset. To improve its performance, a lot of following variants are proposed. DeiT \cite{touvron2021training} is the representative variant of ViT which removes the necessity to pre-train visual transformers on very large scale dataset by inheriting inductive bias from CNNs using knowledge distillation. Despite of the high performance, large computational cost and model size are the significant problems in ViT and its variants.
%PVT~\cite{wang2021pyramid}, on the other hand, proposes a pyramid encoder-decoder architecture with visual transformer as a backbone to perform image recognition, object detection. Different from ViT and DeiT, PVT changes the size of the tokens in encoder and decoder blocks. Some other proceeding studies in this direction can be listed as [cite]. In our study, we use ViT and DeiT models to perform compression on them, however, our method can be applied to other visual transformers as long as they have self-attention heads and MLP layers. 

\textbf{Structured Pruning.} Model compression is a promising strategy to reduce the deployment costs of neural networks. Among various model compression techniques, structured pruning is a very popular choice because its hardware-friendly nature can bring practical efficiency on the real-world devices. Based on different pruning criterias, various structured pruning approaches have been extensively studied in the existing literature \cite{yu2018nisp,zhuang2018discrimination,liu2019metapruning,he2019filter,lin2020hrank,tiwari2021chipnet,lou2022lite}, and most of them focus on pruning CNN models; while the efficient structured pruning of ViTs is little explored. One of these studies,~\cite{chen2021chasing}, prunes the vision transformers using structured pruning.~\cite{yu2021unified}, on the other hand, focuses on FLOPs reduction with the vision transformers using pruning, layer skipping, and knowledge distillation whereas in our study we focus on structured pruning to mainly reduce the number of parameters for building hardware-friendly compressed models. For this reason, we compare our method to~\cite{chen2021chasing}.

% ~\burak{Should we add the arxiv works on structured pruning on ViT?}

% Structured pruning removes regular patterns such as columns of weight matrix, convolutional filters and channels. Structured pruning works usually remove structured patterns according to group lasso \cite{li2016pruning}, geometric median \cite{he2019filter}, ranks of feature maps \cite{lin2020hrank} etc. Even these pruning strategies provide high compression ratio without significant performance drop, they are still based on CNNs and pruning on vision transformer remains unexplored.

% For unstructured pruning, weight pruning is the representative one that removes some entries of weight in each layer. Early works \cite{han2015deep, han2015learning}, either directly prune and retrain the model  or introduce norm regularization terms into the training loss function. Recent weight pruning works prune the least important entries according to multiple criterion such as second order information \cite{NIPS2017_c5dc3e08}, trainable importance mask \cite{huang2018data}, sparsity constraint \cite{liu2017learning}, explainable back-propagation score \cite{yu2018nisp} etc. The heaviest problem of unstructured pruning is that it is very difficult to deploy sparse DNN model without structure on hardware platform due to the irregular memory access and the extra need of weight indexing. 

% \textbf{Sparse Vision Transformers.} XXXXXX

% related works limitations:
% 1) unstructured
% 2) criterion for pruning
%   head: Taylor
%   MLP:  l1 norm

\section{Structured Pruning of ViTs: Analysis}
\label{sec:analysis}
% In this section, we firstly introduce the basic concept of structured pruning on vision transformer. Then we explain the key idea of the proposed approach based on Markov Chain.
\textbf{Notation.} Considering an $L$-block vision transformer, $\bm{W}_{\mathrm{attn}}^{(l)}=\{\bm{W}_{\mathrm{qkv}}^{(l)}, \bm{W}_{\mathrm{proj}}^{(l)}\}$ and $\bm{W}_{\mathrm{mlp}}^{(l)}=\{\bm{W}_{\mathrm{fc1}}^{(l)}, \bm{W}_{\mathrm{fc2}}^{(l)}\}$ represent the weights of the attention layer and the MLP layer at $l$-th block, respectively. For each attention layer, there are $H$ self-attention heads, namely $\bm{W}_{\mathrm{qkv}}^{(l)}=\{\bm{W}_{\mathrm{qkv}}^{(l,h)}\}_{h=1}^H$ and $\bm{W}_{\mathrm{proj}}^{(l)}=\{\bm{W}_{\mathrm{proj}}^{(l,h)}\}_{h=1}^H$. To simplify the notation, in the following content we take one block as the example and omit the superscript (layer index). 
\begin{figure}[bt]
    \centering
    \includegraphics[width=0.7\linewidth]{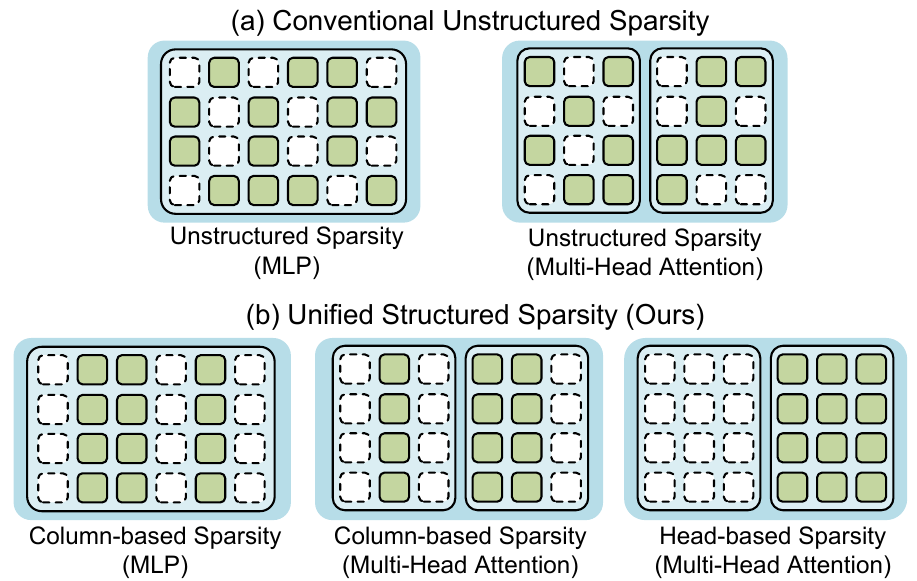}
    \caption{(a) Sparsity pattern of ViT models after unstructured pruning. Only part of the Multi-Head Attention and MLP columns are pruned which are not hardware-friendly.(b) Heterogeneous sparsity patterns of ViT models after structured pruning. Certain MLP and Multi-Head Attention columns are removed which is hardware-friendly.}
    \label{fig:group_norm}
\vspace{-1.7em}
\end{figure}

\textbf{Heterogeneity of structured sparsity.} Because of the difference of the network architecture, the meaning of `structured sparsity' varies with different model types. As described and performed in \cite{wen2016learning,anwar2017structured,liu2018rethinking,liu2020autocompress}, the structured pruning of CNN and RNN typically indicates the removal of the entire channels of the weight tensors and the entire columns of the weight matrices, respectively. Notice that here for either of these two cases, only one type of the structured sparse pattern exist because of the \textit{architectural homogeneity} of the CNN and RNN.

On the other hand, a ViT model exhibits inherent \textit{architectural heterogeneity}. Within the same block, the front-end multi-head attention module and the back-end MLP module represent two types of design philosophy for information processing, and thereby leading to huge difference on both computing procedures and the available structured sparse patterns.

To be specific, when we consider performing structured pruning of ViT model, three types of structured sparse patterns can co-exist with different levels of granularity across different modules. For the multi-head attention module, because each attention head is processing the information individually in a parallel way, the pruning can be performed at the \textit{head-level} to sparsify this component. In addition, consider the weights in the heads are represented in the matrix format; the \textit{column-level} sparsity can also be introduced towards structured pruning. Meanwhile, because the MLP consists of multiple weight matrices as well, the column-level of granularity sparsity can be imposed on this back-end module at the same time. Consequently, a structured pruned ViT model can exhibit heterogeneous structured sparsity (see Fig. \ref{fig:group_norm}). 

\textbf{Problem Definition.} Based on the above analysis, the structured pruning of a vision transformer model with loss function $\ell(\cdot)$ can be formulated as the following general optimization problem:
\begin{equation}
\label{eq:org}
\begin{aligned}
\min_{\bm{W}_{\mathrm{attn}}, \bm{W}_{\mathrm{mlp}}}& \ell(\bm{W}_{\mathrm{attn}}, \bm{W}_{\mathrm{mlp}}),\\
\textrm{s.t.}~~~~~~~&\|\bm{W}_{\mathrm{attn}}\|_{0}^{\mathrm{h}}\le \kappa_{\mathrm{attn}}^{\mathrm{h}},\\
&\|\bm{W}_{\mathrm{attn}}\|_{0}^{\mathrm{c}}\le \kappa_{\mathrm{attn}}^{\mathrm{c}},\\
&\|\bm{W}_{\mathrm{mlp}}\|_{0}^{\mathrm{c}}\le \kappa_{\mathrm{mlp}}^{\mathrm{c}},
\end{aligned}
\end{equation}
where $\kappa^{\mathrm{c}}$ and $\kappa^{\mathrm{h}}$ are the desired number of columns and the desired number of heads after pruning, respectively. $\|\cdot\|_0^{\mathrm{c}}$ and $\|\cdot\|_0^{\mathrm{h}}$ are the column-based and head-based group $L_0$-norm, which denote the number of non-zero columns and the number of non-zero heads, respectively.

% To optimize the problem with $L_0$-norm is generally NP-hard, thus conventional pruning methods directly prune the heads and neurons with least $L_0$-norm values from a pre-trained model by one shot, then retrain the pruned model.
\textbf{Questions to be Answered.} Solving the above optimization problem is non-trivial since it contains the constraints involved with multi-granularity sparsity for different model components. More specifically, two important questions need to be answered.
\begin{figure*}[t!]
    \centering
    \includegraphics[width=\linewidth]{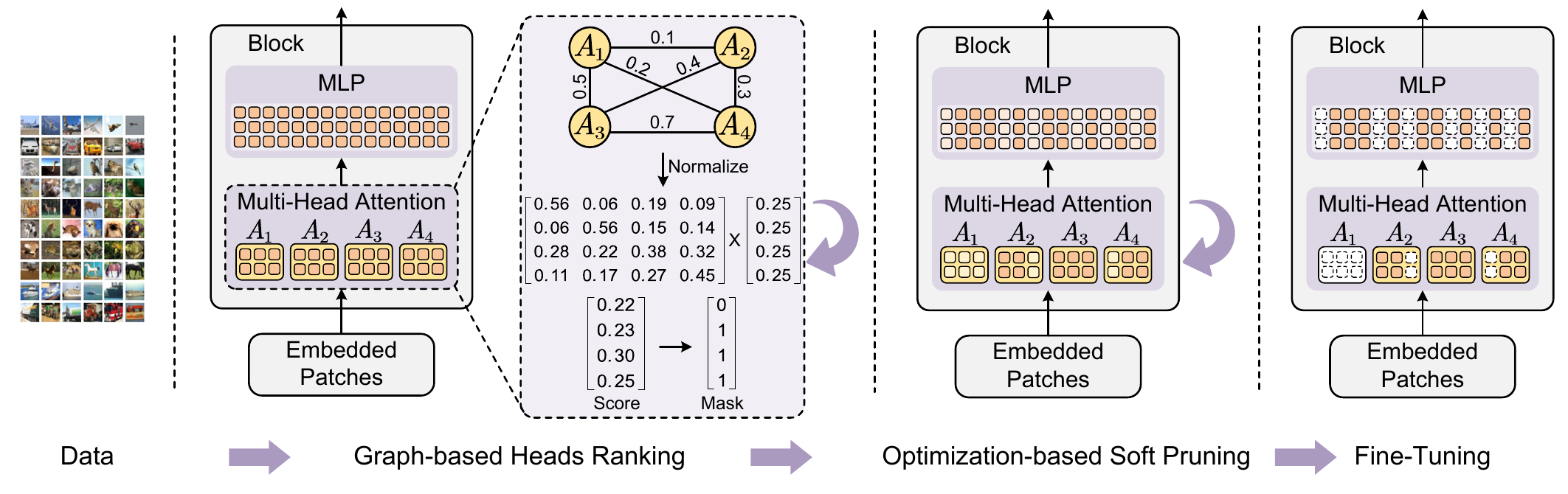}
    \vspace{-7mm}
    \caption{Procedure of the proposed multi-stage structured pruning approach. }
    \label{fig:arch}
\end{figure*}
\underline{\textbf{Question \#1:}} \textit{What is the suitable pruning criterion to obtain head-level sparsity?}

\underline{Analysis:} From the perspective of information processing, multi-head attention shares some interesting similarity with convolutional layer. Both of them use multiple individual computing units, i.e., attention heads and convolutional filters, to perform parallel computations. Therefore, a naive way to perform head-level pruning is to leverage the existing criteria developed in the channel pruning of CNNs. However, such straightforward solution, in principle, may not be the best choice because of two reasons. First, the receptive fields and the focused locality of the attention head and filters are different, and hence simply using the criterion for pruning channels is not a suitable strategy. Second and more importantly, most of the existing channel pruning criterias are built on the information of each individual channel (the corresponding filter weight and/or its feature map). When adopting this philosophy in the head pruning, the insufficient utilization of inter-head information will probably cause non-negligible performance loss. Overall, the unique characteristics of multi-head attention mechanism calls for attention-specific pruning criterion.

\underline{\textbf{Question \#2:}} \textit{How should we coordinate the pruning across different modules with different levels of granularity? }

\underline{Analysis:} As indicated before, three types of structured sparse pattern can co-exist in the different modules of the pruned ViT models. A key component of the to-be-explored structured pruning strategy is to develop a good coordination scheme that can properly impose these different structured sparse patterns in a joint and global way. Consider the complicated interaction among different types of structured sparsity, the expected pruning strategy should be able to solve this problem in a systematic and global way.

\section{Structured Pruning of ViTs: Method}

\subsection{Graph-based Head Ranking}

To answer Question \#1, we propose a graph-based approach to measure and determine the importance of different attention heads, which can be further used for the follow-up pruning. \textbf{Our key idea} is to model the inter-head correlation as a graph, and then leverage the \textit{graph-based ranking}, a methodology that has been successfully used in many web search and NLP algorithms, such as PageRank \cite{page1999pagerank}, TextRank \cite{mihalcea2004textrank} and LexRank \cite{erkan2004lexrank}, to select important attention heads. 

\textbf{Graph Construction of Markov Chain.} To be specific, we first construct a graph $G=(\bm{A},\bm{E})$ to represent the  attention heads and their similarities in the block of a ViT model. The set of nodes $\bm{A}$ denote all the attention heads $\{A_h\}_{h=1}^H$, and $\bm{E}$ is the set of connected edges. For edge $E(A_i, A_j)$, its weight is defined as the expected cosine similarity between $A_i$ and $A_j$. According to \cite{mihalcea2004textrank}, the graph defined with such cosine similarity can be interpreted as a \textit{Markov chain}, where each node is a state, and the transition probability $P(i,j)$ between two states is the edge weight. In such scenario, $P(i,j)$ can be calculated as: 
\begin{equation}
\label{eq:expect}
P(i,j)=\mathbb{E}_{\bm{X}\sim\mathcal{D}}\left[\texttt{CosineSim}(A_i(\bm{X}), A_j(\bm{X}))\right],
\end{equation}
where $A_i(\bm{X})$ is the output of $i$-th attention head with sampled input $\bm{X}$ and $\mathcal{D}$ is the data set. Built upon this calculation, the entire transition matrix $\bm{P}$ of a Markov chain. Notice that as indicated in  \cite{erkan2004lexrank}, each column of $\bm{P}$ should be further normalized.

\underline{Batch estimation.} Calculating the transition probability can be very costly since it needs to be performed across the entire training dataset $\mathcal{D}$ (see Eq. \ref{eq:expect}). To solve this problem, we adopt a batch-based estimation strategy to improve computation efficiency without sacrificing ranking performance. To be specific, as described in Eq. \ref{eqn:cosine}, only a batch of training data is sampled and used to to estimate the transition probability. As our ablation study in Section~\ref{sect:ablation} will show, using different batch sizes ($B$) bring very stable ranking results for the attention heads, thereby empirically verifying the effectiveness of this estimation strategy.
\begin{equation}
P(i,j)=\texttt{CosineSim}\left(\sum_{b=1}^{B}A_i(\bm{X}_b),\sum_{b=1}^{B}A_j(\bm{X}_b)\right).
\label{eqn:cosine}
\end{equation}
% \textbf{Markov Chain.} \cite{gagniuc2017markov} A Markov chain is a type of Markov process, which satisfies the property that the future state is only conditional on the current state and independent from the past states. The \textit{transition matrix} describes the probabilities associated with state changes from one state to the others across state space. Given the transition matrix and initial state, a Markov chain is generally defined. Markov chain is already efficiently used to solve ranking problem for decades, e.g. \textit{PageRank} \cite{page1999pagerank}, the backbone of Google search engine, is one of the most famous applications. However, this elegant ranking algorithm is still never investigated in DNN architecture optimization.

\textbf{Importance Ranking.} Mathematically, an irreducible and aperiodic Markov chain is guaranteed to converge to a stationary distribution \cite{seneta2006non}. As indicated in \cite{erkan2004lexrank}, once converged, the probability of a random walker stays in one state can reflect the state importance. Motivated by this observation, we propose to quantify the importance of each attention head via calculating the stationary distribution in our constructed Markov chain. To that end, the iterative \textit{power method} \cite{erkan2004lexrank} can be used via setting a uniform distribution for the states as the initialization. Overall, the entire graph-based head ranking procedure is described in Algorithm \ref{alg:ranking}.
\begin{algorithm}[!b]
\caption{Graph-based Attention Head Ranking}
\label{alg:ranking}
\textbf{Input}: Sampled batch $\{\bm{X}_b\}_{b=1}^B$, attention heads $\{A_h\}_{h=1}^H$;\\
% \textbf{Parameter}: Optional list of parameters\\
\textbf{Output}: Importance score $\bm{s}=[s_1,\cdots,s_H]$. \par
\begin{algorithmic}[1] %[1] enables line numbers
\STATE Initialize transition matrix: $\bm{P}:=\texttt{zeros}(H,H)$;
\FOR{$i=1$ to $H$}
  \FOR{$j=1$ to $H$}
    \STATE Calculate $P(i,j)$ via Eq. \ref{eqn:cosine};
  \ENDFOR
\ENDFOR
\STATE Normalize each column of $\bm{P}$;
\STATE Initialize $\bm{s}:=\texttt{ones}(H)/H$;
\REPEAT 
\STATE $\bm{s}':=\bm{s}$;
\STATE $\bm{s}:=\bm{P}s$;
\STATE $\delta:=\|\bm{s}-\bm{s}'\|$;
\UNTIL $\delta\le\epsilon$
% \STATE Calculate score $\bm{s}$ via Alg. \ref{alg:power} with $\bm{P}$.
% \STATE \textbf{return} solution
\end{algorithmic}
\end{algorithm}
% Let vector $\bm{s}$ with size $H$ represent the state distribution of the ranking process. According to the properties of Markov chain, the next state only depends on the previous state, thus the state distribution at time $t$ can be calculated by
% \begin{equation}
%     \bm{s}^{t} = \bm{P}\bm{s}^{t-1}.
%     \label{eqn:state_update}
% \end{equation}

% Intuitively, the stationary distribution can be interpreted by \textit{random walk}. A random walker will probably step to the next node with larger similarity. Finally, the walker get stuck in the state and no longer visit other state of the graph \cite{erkan2004lexrank}.

% In other words, the stationary distribution indicates the probabilities that the random walker finally stay at which node. We can consider this stationary distribution as the importance scores to rank attention heads.

\textbf{Soft-Pruning Mask.} Once the importance score for each state is obtained via calculating the stationary distribution, the corresponding attention heads can be ranked. Here we use a binary mark matrix $\bm{M}_{\mathrm{attn}}=\{\bm{M}_{\mathrm{qkv}}, \bm{M}_{\mathrm{proj}}\}$ to indicate the weight entries associated with the least important heads that should be removed. Notice that at this stage the head pruning is not performed yet. Instead such binary mask serves as the guideline for the next-stage optimization, and it is essentially integrated into Eq. \ref{eq:org} as follows: 
\begin{equation}
\label{eqn:obj_1}
\begin{aligned}
\min_{\bm{W}_{\mathrm{attn}}, \bm{W}_{\mathrm{mlp}}}& \ell(\bm{W}_{\mathrm{attn}}, \bm{W}_{\mathrm{mlp}}) \\
\textrm{s.t.}~~~~~~~&\|(\bm{1}-\bm{M}_{\mathrm{attn}})\odot\bm{W}_{\mathrm{attn}}\|_{0}= 0,\\
&\|\bm{W}_{\mathrm{mlp}}\|_{0}\le \kappa_{\mathrm{mlp}}^{\mathrm{c}},\\
&\|\bm{M}_{\mathrm{attn}}\odot\bm{W}_{\mathrm{attn}}\|_{0}^{\mathrm{c}}\le \kappa_{\mathrm{attn}}^{\mathrm{c}},
\end{aligned}
\end{equation}
where $\odot$ is element-wise product. In general, because the overall optimization phase coordinates and adjusts the different types of structured sparse pattern from a global perspective, this ranking-only "soft" pruning strategy, instead of directly pruning the least important heads, can provide more flexibility and possibility for the next-stage optimization procedure to identify better structured sparse models. 
% \begin{algorithm}[tb]
% \caption{Power Method}
% \label{alg:power}
% \textbf{Input}: Transition matrix $\bm{P}$, tolerance $\epsilon$, number of state $H$;\\
% % \textbf{Parameter}: Optional list of parameters\\
% \textbf{Output}: Stationary state distribution $\bm{s}$. \par
% \begin{algorithmic}[1] %[1] enables line numbers
% \STATE Initialize $\bm{s}:=\texttt{ones}(H)/H$;
% \REPEAT
% \STATE $\bm{s}':=\bm{s}$;
% \STATE $\bm{s}:=\bm{P}s$;
% \STATE $\delta:=\|\bm{s}-\bm{s}'\|$;
% \UNTIL $\delta\le\epsilon$
% % \STATE \textbf{return} solution
% \end{algorithmic}
% \end{algorithm}
\vspace{-1.7em}
\subsection{Optimization-based Structured Pruning}
\label{sect:optimization}
As pointed out by Question \#2, the co-existence of multi-granularity and multi-location of the sparsity of ViT models make the entire structured pruning procedure become very challenging. To solve this, we propose to use advanced optimization technique to perform systematic structured pruning. To be specific, considering the complicated interactions among different types of structured sparsity, we do not prune the heads or columns immediately, since any direct hard pruning at the early stage may cause severe accuracy loss. Instead, we adopt "soft-pruning" strategy via optimizing the entire ViT models towards the desired structured sparse formats. In other words, the three types of sparsity pattern are gradually imposed onto the attention heads and MLPs.

To that end, we first relax the constraints of Eq. \ref{eqn:obj_1} and rewrite it as follows:
\begin{equation}
\label{eqn:obj_2}
\begin{aligned}
\min_{\bm{W}_{\mathrm{attn}}, \bm{W}_{\mathrm{mlp}}}& \ell(\bm{W}_{\mathrm{attn}}, \bm{W}_{\mathrm{mlp}}) +\frac{\lambda}{2}\|(\bm{1}-\bm{M}_{\mathrm{attn}})\odot\bm{W}_{\mathrm{attn}}\|_{F}^2,\\
\textrm{s.t.}~~~~&\|\bm{W}_{\mathrm{mlp}}\|_{0}^{\mathrm{c}}\le \kappa_{\mathrm{mlp}}^{\mathrm{c}},\\
&\|\bm{M}_{\mathrm{attn}}\odot\bm{W}_{\mathrm{attn}}\|_{0}^{\mathrm{c}}\le \kappa_{\mathrm{attn}}^{\mathrm{c}},
\end{aligned}
\end{equation}
where $\lambda$ is the coefficient that controls the influence of quadratic term. 

\begin{algorithm}[tb]
\caption{Overall Procedure of GOHSP Framework}
\label{alg:overall}
\textbf{Input}: Dense weight $\{\bm{W}_{\mathrm{attn}}, \bm{W}_{\mathrm{mlp}}\}$, desired model size $\{\kappa_{\mathrm{attn}},\kappa_{\mathrm{mlp}}\}$, training data $\mathcal{D}$, number of epochs $E$;\\
\textbf{Output}: Structured sparse weight $\{\tilde{\bm{W}}_{\mathrm{attn}}, \tilde{\bm{W}}_{\mathrm{mlp}}\}$;\par
\begin{algorithmic}[1] %[1] enables line numbers
\STATE Sample a batch of data $\{X_b\}_{b=1}^{B}$ from $\mathcal{D}$;
\STATE Calculate importance score $\bm{s}$ via Alg. \ref{alg:ranking};
\STATE Obtain structured mask $\bm{M}_{\mathrm{attn}}$ according to $\bm{s}$;
\STATE $\bm{Z}_{\mathrm{attn}}:=\bm{W}_{\mathrm{attn}}$, $\bm{Z}_{\mathrm{mlp}}:=\bm{W}_{\mathrm{mlp}}$; \textcolor{ForestGreen}{\textit{// Initialize auxiliary variables}}
\STATE $\bm{U}_{\mathrm{attn}}:=\bm{0}$, $\bm{U}_{\mathrm{mlp}}:=\bm{0}$; \textcolor{ForestGreen}{\textit{// Initialize Lagrangian multipliers}}
\FOR{$e=1$ to $E$}
\STATE Update $\bm{W}_{\mathrm{attn}},\bm{W}_{\mathrm{attn}}$ via Eq. \ref{eqn:update_w_attn} and Eq. \ref{eqn:update_w_mlp};
\STATE Update $\bm{Z}_{\mathrm{attn}},\bm{Z}_{\mathrm{mlp}}$ via Eq. \ref{eqn:update_z_attn} and Eq. \ref{eqn:update_z_mlp};
\STATE Update $\bm{U}_{\mathrm{attn}}, \bm{U}_{\mathrm{mlp}}$ via Eq. \ref{eqn:update_u_attn} and Eq. \ref{eqn:update_u_mlp};
\ENDFOR
\STATE Fine-tune pruned weight $\{\tilde{\bm{W}}_{\mathrm{attn}}, \tilde{\bm{W}}_{\mathrm{mlp}}\}$.
\end{algorithmic}
\end{algorithm}

\textbf{Optimization-based Soft Pruning.} As indicated in \cite{boyd2011distributed}, when the constraints of continuous non-convex problem are sparsity related (as Eq. \ref{eqn:obj_2} shows), Douglas—Rachford splitting method \cite{eckstein1992douglas} can be a suitable optimization solution for such types of problem. Following this philosophy, we first introduce auxiliary variables $\bm{Z}_{\mathrm{attn}},\bm{Z}_{\mathrm{mlp}}$ and indicator functions as:
\begin{align}
&g(\bm{Z}_{\mathrm{attn}})=
\begin{cases}
0 & \|\bm{M}_{\mathrm{attn}}\odot\bm{Z}_{\mathrm{attn}}\|_{0}^{\mathrm{c}}\le \kappa_{\mathrm{attn}}^{\mathrm{c}},\\ 
+\infty & \textrm{otherwise},
\end{cases}\\
&h(\bm{Z}_{\mathrm{mlp}})=
\begin{cases}
0 & \|\bm{Z}_{\mathrm{mlp}}\|_{0}^{\mathrm{c}}\le \kappa_{\mathrm{mlp}}^{\mathrm{c}},\\
+\infty & \textrm{otherwise}.
\end{cases}
\end{align}
Then, we can rewrite Eq. \ref{eqn:obj_2} as the following equivalent form:
\begin{equation}
\begin{aligned}
\min_{\bm{W}, \bm{Z}}~~~ \ell(\bm{W}_{\mathrm{attn}}, \bm{W}_{\mathrm{mlp}})&+g(\bm{Z}_{\mathrm{attn}})+
h(\bm{Z}_{\mathrm{mlp}}) +\\ \frac{\lambda}{2}\|(\bm{1}-\bm{M}_{\mathrm{attn}})&\odot\bm{W}_{\mathrm{attn}}\|_{F}^2,\\
\textrm{s.t.}~~~~&\bm{W}_{\mathrm{mlp}} = \bm{Z}_{\mathrm{mlp}},\\
&\bm{W}_{\mathrm{attn}} = \bm{Z}_{\mathrm{attn}}.
\end{aligned}
\label{eqn:obj_3}
\end{equation}
In such scenario, the corresponding augmented Lagrangian function of the above optimization objective is:
\begin{equation}
\begin{aligned}
\mathcal{L}_{\rho}(\bm{W}_{\mathrm{attn}}, \bm{W}_{\mathrm{mlp}},\bm{Z}_{\mathrm{mlp}})=\ell({\bm{W}_{\mathrm{attn}}, \bm{W}_{\mathrm{mlp}}})+g(\bm{Z}_{\mathrm{attn}})+&\\ h(\bm{Z}_{\mathrm{mlp}})+
\frac{\lambda}{2}\|(\bm{1}-\bm{M}_{\mathrm{attn}})\odot\bm{W}_{\mathrm{attn}}\|_{F}^2+&\\
\frac{\rho}{2}\|\bm{W}_{\mathrm{attn}}-\bm{Z}_{\mathrm{attn}}+\bm{U}_{\mathrm{attn}}\|_F^2+&\\
\frac{\rho}{2}\|\bm{U}_{\mathrm{attn}}\|_F^2+
\frac{\rho}{2}\|\bm{W}_{\mathrm{mlp}}-\bm{Z}_{\mathrm{mlp}}+\bm{U}_{\mathrm{mlp}}\|_F^2+\frac{\rho}{2}\|\bm{U}_{\mathrm{mlp}}\|_F^2,
\end{aligned}
\label{eqn:lagrangian}
\end{equation}
where $\rho>0$ is the penalty parameter, and $\bm{U}_{\mathrm{attn}},\bm{U}_{\mathrm{mlp}}$ are the Lagrangian multipliers. Then the variables at step $t$ can be iteratively updated as:
\begin{equation}
\begin{aligned}
\bm{W}_{\mathrm{attn}}^{t}=\bm{W}_{\mathrm{attn}}^{t-1}-&\eta\frac{\ell({\bm{W}_{\mathrm{attn}}, \bm{W}_{\mathrm{mlp}}^{t-1}})}{\bm{W}_{\mathrm{attn}}}-\\
\lambda\left[(\bm{1}-\bm{M}_{\mathrm{attn}})\odot\bm{W}_{\mathrm{attn}}^{t-1}\right]-
&\rho(\bm{W}_{\mathrm{attn}}^{t-1}-\bm{Z}_{\mathrm{attn}}^{t-1}+\bm{U}_{\mathrm{attn}}^{t-1}),
\end{aligned}
\label{eqn:update_w_attn}
\end{equation}
\begin{equation}
\begin{aligned}
\bm{W}_{\mathrm{mlp}}^{t}=\bm{W}_{\mathrm{mlp}}^{t-1}-\eta\frac{\ell({\bm{W}_{\mathrm{attn}}^{t}, \bm{W}_{\mathrm{mlp}}})}{\bm{W}_{\mathrm{mlp}}}-&\\
\rho(\bm{W}_{\mathrm{mlp}}^{t-1}-\bm{Z}_{\mathrm{mlp}}^{t-1}+\bm{U}_{\mathrm{mlp}}^{t-1}),~
\end{aligned}
\label{eqn:update_w_mlp}
\end{equation}
\begin{equation}
\bm{Z}_{\mathrm{attn}}^{t}=\bm{\mathcal{P}}(\bm{W}_{\mathrm{attn}}^{t}+\bm{U}_{\mathrm{attn}}^{t-1}),
\label{eqn:update_z_attn}
\end{equation}
\begin{equation}
\bm{Z}_{\mathrm{mlp}}^{t}=\bm{\mathcal{P}}(\bm{W}_{\mathrm{mlp}}^{t}+\bm{U}_{\mathrm{mlp}}^{t-1}),
\label{eqn:update_z_mlp}
\end{equation}
\begin{equation}
\bm{U}_{\mathrm{attn}}^{t}=\bm{U}_{\mathrm{attn}}^{t-1}+\bm{W}_{\mathrm{attn}}^{t}-\bm{Z}_{\mathrm{attn}}^{t},
\label{eqn:update_u_attn}
\end{equation}
\begin{equation}
\bm{U}_{\mathrm{mlp}}^{t}=\bm{U}_{\mathrm{mlp}}^{t-1}+\bm{W}_{\mathrm{mlp}}^{t}-\bm{Z}_{\mathrm{mlp}}^{t}.
\label{eqn:update_u_mlp}
\end{equation}
Here $\eta$ is the optimizer learning rate for training the ViT, and $\bm{\mathcal{P}}$ is the Euclidean projection for the sparse constraint.

\textbf{Final Hard-Pruning and Fine-Tuning.} After the above described optimization procedure, the structured sparse patterns have been gradually imposed onto the ViT model. In other words, the weight values of the masked attention heads, as well as some columns of MLPs and attention heads, become extremely small. At this stage, we can now prune those small weights and then perform a few rounds of fine-tuning to achieve higher performance.

Overall, by using graph-based head ranking and optimization-based structured pruning, the previously raised Question \#1 and \#2 can be properly addressed. The overall GOHSP framework is summarized in Fig. \ref{fig:arch}.
% In overall, the proposed approach contains three stages. At the first stage, we construct a graph whose nodes are all the attention heads. In the graph, the probability transiting from one node to another is calculated by cosine similarity between the outputs of the corresponding attention heads. As random walk in the constructed graph, the heads ranking can be considered as Markov Chain, where the stationary state distribution is the ranking score of each node. Then, we can obtain a mask by marking the self attention heads with least score as removed. At the second stage, instead of hard pruning, we gradually prune the least important attention heads marked in the previous stage during training. To achieve this, we formulate an optimization objectives with structured sparsity constraints, then we propose to solve this problem using advanced optimization algorithm. After training with optimization-based soft pruning, the weight values associated with the masked attention heads become extremely small such that these values can be regarded as non-influential on the prediction of the vision transformer model. At the third stage, we direct remove these attention heads with negligible values, then we can obtained a well compressed model with high performance by fine-tuning the pruned model with masked gradients for only few epochs. The overall algorithm procedure is summarized in Alg. \ref{alg:overall}. 
\section{Experiments}
\begin{table}[!b]
\vspace{-1mm}
\caption{Performance comparison between our GOHSP and structured one-shot/gradually magnitude-based pruning (SOMP/SGMP) of ViT-Small model on CIFAR-10 dataset. 
%Our model improves the accuracy of the baseline slightly while reducing the number of parameters by 40\%.
}
\begin{center}
\resizebox{0.45\textwidth}{!}{
\begin{tabular}{lccc}
\toprule
\textbf{Method} & \textbf{Sparsity}  & \textbf{\# Paramters} & \textbf{Top-1 (\%)}\\
\midrule
Baseline & - & 48.0M & 97.85\\
\midrule
SOMP &  40\% & 28.8M & 96.07\\
SGMP & 40\% & 28.8M &96.93\\
\textbf{GOHSP (Ours)} & 40\% & 28.8M & \textbf{97.89}\\
\textbf{GOHSP (Ours)} & 80\% & 9.6M & 97.40\\
\bottomrule
\end{tabular}}
\end{center}
\label{tbl:cifar10}
\end{table}
\begin{table*}[!t]
\caption{Comparison results of our method, GOHSP, with other structured and unstructured pruning methods on ImageNet.}
\vspace{-1mm}
\begin{center}
\resizebox{0.75\textwidth}{!}{
\begin{tabular}{llccccc}
\toprule
\textbf{Model} & \textbf{Method} & \textbf{Sparsity} & \textbf{\# Parameters} & \textbf{FLOPs $\downarrow$} & \textbf{Run-time $\downarrow$} & \textbf{Top-1 (\%)}\\
\midrule
\multirow{7}{*}{\textbf{DeiT-Tiny}} & Baseline & - & 5.7M & - & - &72.20\\
\cmidrule{2-7}
& OMP (Unstructured) & 30\% & 4.02M & 25.56\% & -&68.35\\
& GMP (Unstructured) & 30\% & 4.02M & 25.56\% &- &69.56\\
& TP (Unstructured) & 30\% & 4.02M & 25.56\% & -&68.38\\
& SSP (Structured) & 30\% & 4.2M & 23.69\% & -&68.59\\
& S$^2$ViTE (Structured) & 30\% & 4.2M & 23.69\% & 10.57 \% &70.12\\
& \textbf{GOHSP} (Structured) & 30\% & \textbf{4.0M} & \textbf{30\%} & \textbf{13.41}\% & \textbf{70.24}\\
\midrule
\multirow{4}{*}{\textbf{DeiT-Small}} & Baseline & - & 22.1M & - & - & 79.90\\
\cmidrule{2-7}
% & OMP (Unstructured) & 50\% & 11.1M & 46.26\% &76.32\\
% & GMP (Unstructured) & 50\% & 11.1M & 46.26\% &76.88\\
% & TP (Unstructured) & 50\% & 11.1M & 46.26\% &76.30\\
& SSP (Structured) & 40\% & 14.6M & 31.63\% & - &77.74\\
& S$^2$ViTE (Structured) & 40\% & 14.6M & 31.63\% & 22.65\% &79.22\\
% & \textbf{GOHSP} (Structured) & 50\% & \textbf{11.1M} & \textbf{48\%} & \\
& \textbf{GOHSP} (Structured) & 40\% & 14.4M & 35\% & 24.61\% & \textbf{79.98}\\
& \textbf{GOHSP} (Structured) & 50\% & 11.1M & 39\% & \textbf{26.57}\% & 79.86\\
\bottomrule
\end{tabular}}
\end{center}
\label{tbl:imagenet}
\end{table*}
% \miao{Experiment list:\\
% -- DeiT-Tiny on ImageNet (done)\\
% -- layer-wise distillation for DeiT-Small on ImageNet (done for ViT-Small on CIFAR-10, improves 0.6 point 97.9\%) (running)\\
% -- CIFAR-10 (done)\\
% -- Performance curves w.r.t. different sparsities, ours vs. hard pruning (done)\\
% -- Our ranking vs. random selection (running)\\
% -- training curves w.r.t. different $\rho$ (done)\\
% -- visualization of ranking results w.r.t. different batch sizes (done)\\
% -- visualization of ranking results w.r.t. sampled batches
% }
\subsection{Experimental Settings}
\textbf{Dataset and Baseline.} We evaluate the performance of our proposed GOHSP approach on CIFAR-10 and ImageNet datasets~\cite{deng2009imagenet}. For experiments on the CIFAR-10 dataset, the original dense model is ViT-Small\footnote{We take this model from open source library timm. }\cite{dosovitskiy2020image} with 48M parameters. For experiments on the ImageNet dataset, the original dense models are DeiT-Tiny and DeiT-Small~\cite{touvron2021training} with 5.7M and 22.1M parameters, respectively.  
%Baselines are explained in \textbf{Appendix}.

\textbf{Hyper-parameters and Sparsity Ratio.} For our experiments on the CIFAR-10 dataset, the batch size, learning rate and $\rho$ are set as 256, 0.1 and 0.001, respectively. For ImageNet dataset, the batch size, learning rate and $\rho$ are set as 256, 0.01 and 0.001, respectively. For both of these two experiments, SGD is selected as the training optimizer without using weight decay, and we apply Erdős-Rényi \cite{mocanu2018scalable} to determine the sparsity distribution of each layer given an overall sparsity ratio. In particular, soft-pruning maintains high accuracy at the high sparsity ratios.

\subsection{Performance Evaluation}
\textbf{CIFAR-10 Dataset.} Table \ref{tbl:cifar10} shows performance comparison on CIFAR-10 dataset between our proposed GOHSP and other structured pruning method (structured one-shot magnitude pruning (SOMP) \cite{han2015deep} and structured gradually magnitude pruning (SGMP) \cite{zhu2017prune}) for ViT-Small model. It is seen that with the same sparsity ratio, our approach brings significant performance improvement. Compared to SGMP approach, GOHSP achieves $0.96\%$ accuracy increase with the same pruned model size. Even compared with the baseline, the structured sparse model pruned by GOHSP can outperform the uncompressed model with $40\%$ fewer parameters while $80\%$ compressed model achieves only $0.45\%$ worse than the full ViT-Small model.

\textbf{ImageNet Dataset.} Table \ref{tbl:imagenet} summarizes the performance on ImageNet dataset between GOHSP and other structured pruning approaches (SOMP, SGMP, Talyer pruning (TP), Salience-based Structured Pruning (SSP) and S$^2$ViTE\cite{chen2021chasing}) for DeiT-Tiny and DeiT-Small models. It is seen that due to the limited redundancy in such small-size model, the existing pruning approaches suffer from more than $2.5\%$ accuracy loss when compressing DeiT-Tiny. Instead, with the even fewer parameters and more FLOPs reduction, our GOHSP approach can achieve at least $0.68\%$ accuracy increase over the unstructured pruning approaches. Compared to the structured pruning approach (SSP), our method enjoys $1.65\%$ accuracy improvement with lower storage cost and computational cost. In addition, when compressing DeiT-Small model, with fewer parameters and more FLOPs reduction, our GOHSP approach can achieve $0.76\%$ accuracy increase as compared to the state-of-the-art structured pruning method S$^2$ViTE~\cite{chen2021chasing} and can even outperform the original DeiT-Small. With $50\%$ pruned DeiT-Small we achieve similar accuracy to the full DeiT-Small. Finally, we report $26.57\%$ improvement in run-time efficiency with our $50\%$ pruned DeiT-Small.
\begin{figure}[h]
    \centering
    \includegraphics[width=0.95\linewidth]{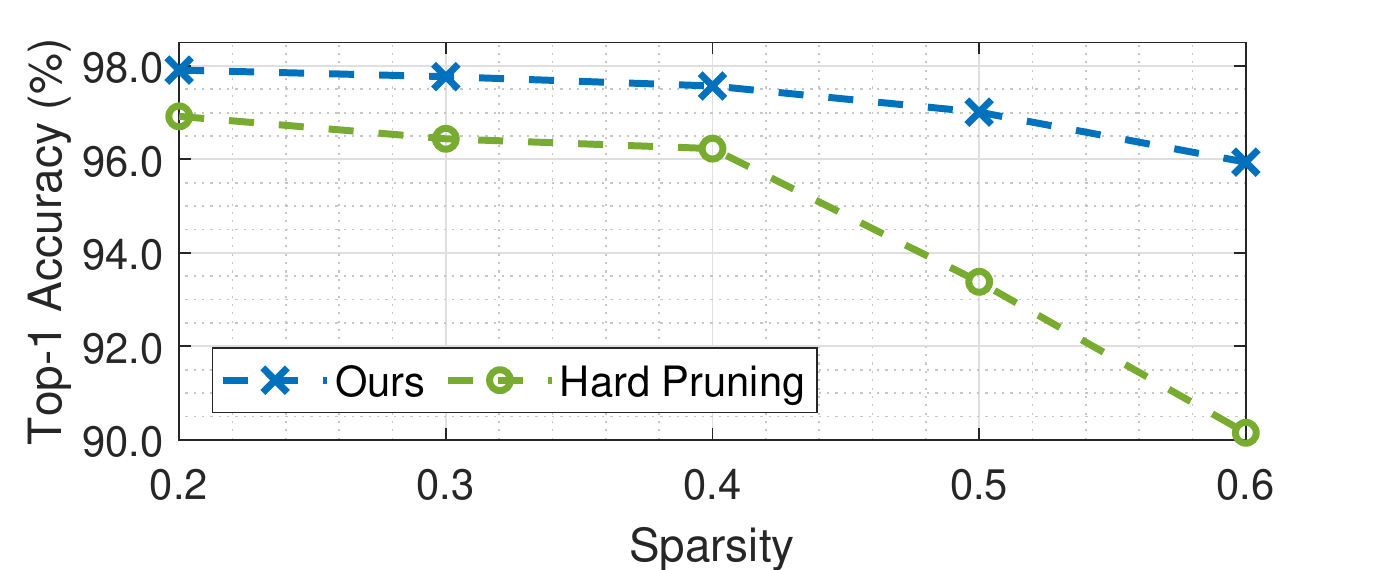}
    \caption{Results on the effect of soft-pruning (ours) and hard-pruning for ViT-Small model on CIFAR-10 dataset.}
    \label{fig:ablation1}
\end{figure}
\begin{figure}[!b]
    \centering
    \includegraphics[width=0.8\linewidth]{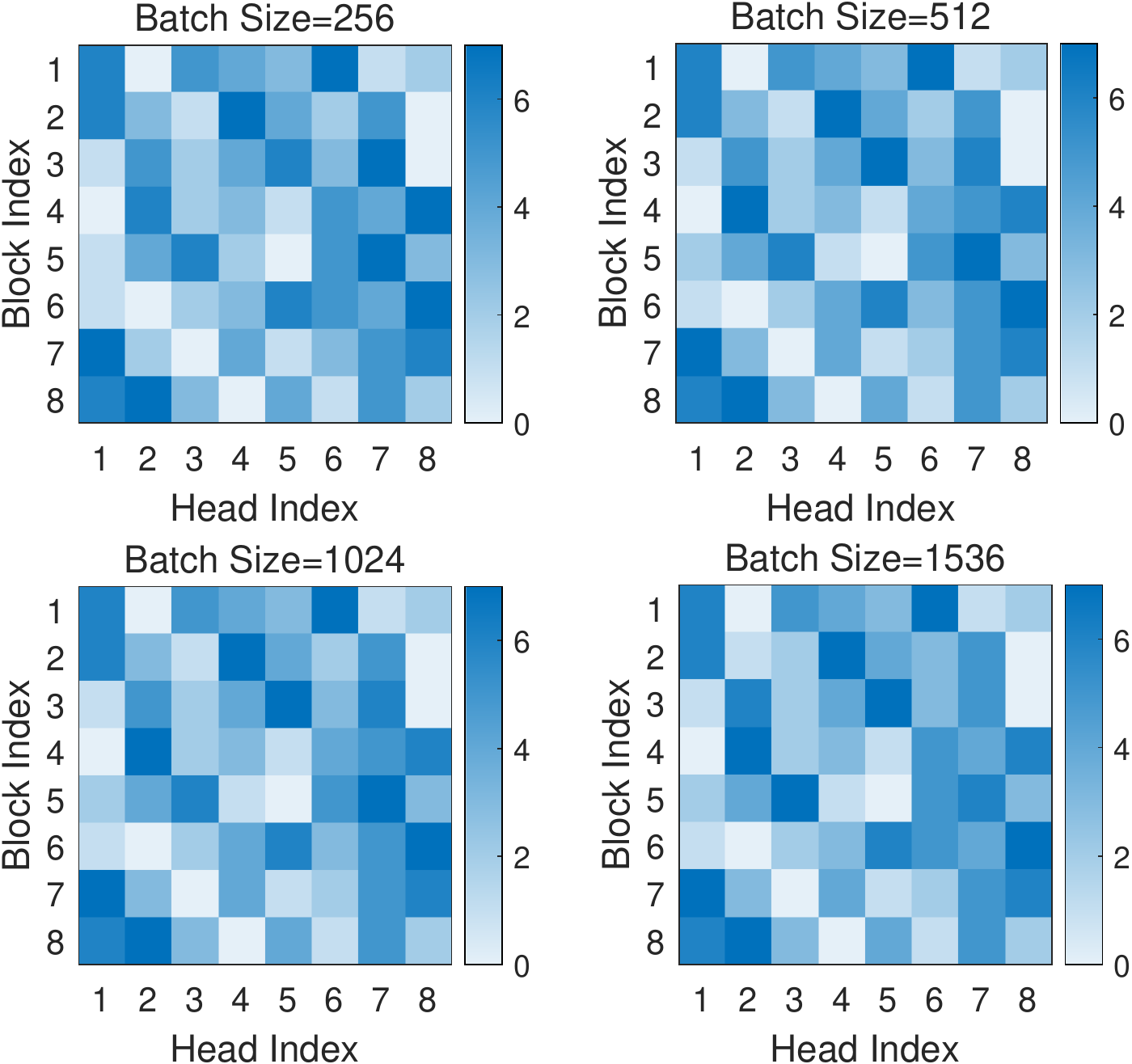}
    \caption{The effect of batch sizes for ranking results. Different colors represent different ranking scores. We can see that our head ranking algorithm is not sensitive to batch size.}
    \label{fig:ablation3}
\end{figure}
% For DeiT-Small, to compare with state-of-the-art unstructured and structured pruning methods, we prune the model with 50\% and 40\% target sparsity, respectively. From Table \ref{tbl:imagenet}, it is seen that the accuracy drop of all the unstructured pruning strategies increases to more than 3\% when compressing DeiT-Small model, while these methods were effective for compressing CNNs. In other words, this results demonstrate these methods are no longer suitable for transformer-based models. On the contrary, with the same sparsity, our method enjoy more than 2\% improvement compared with them. Even though the recent structured sparsity method for vision transformer S$^2$ViTE brings only 0.7\% performance loss, the FLOPs reduction is only 23.7\%. Compared to it, our method enjoys almost no accuracy drop with even more compact model size (our 13.3M vs. S$^2$ViTE 14.6M) and nearly 10\% more FLOPs reduction (ours 32.XX\% vs. S$^2$ViTE 23.7\%).
\subsection{Ablation Study, Visualization and Discussion} \label{sect:ablation}
To obtain the deep understanding of the effect of our proposed approach, we perform several ablation studies and a detailed analysis. Here the experiments conducted in the ablation study focus on compressing ViT-Small on CIFAR-10.

\textbf{Soft Pruning vs Hard Pruning.} As described in Optimization section, after ranking the attention heads, we use the ranking information as a soft-pruning mask to guide the next-phase optimization. The optimization itself is also a soft-pruning procedure that does not directly zero the weights but gradually impose the structured sparsity. To analyze the effect of this strategy, we conduct an ablation experiment via performing the direct hard pruning. In this ablation study, the least important attention heads are removed according to their ranks, and the columns of MLPs with least group $L_1$ norm are also pruned. Such hard pruned models are still trained  with the same hyper-parameters settings that are used for soft pruning method. Fig. \ref{fig:ablation1} shows the curves of top-1 test accuracy with different target sparsity settings. The soft-pruning strategy brings very significant accuracy improvement over the direct hard pruning with the same sparsity ratio.
% \textbf{Our Graph-based Ranking vs. Random Selection.} Besides hard pruning, we also perform ablation study without the graph-based ranking. Instead, we randomly select attention heads to prune satisfying the sparsity rate then preserve the optimization-based soft pruning framework to train the model.
% \burak{They might ask if there is something better than random. We do not want to present LRP, right?}\miao{Here is the ablation study that only demonstrates the effectiveness of our ranking method. We already compare SOTA methods in table 2}
% As shown in Fig. \ref{fig:ablation1}, the performance still drops slightly. This ablation study demonstrates the proposed graph-based ranking is also effective in the proposed approach.
\begin{figure}[!b]
\vspace{-2em}
    \centering
    \includegraphics[width=0.99\linewidth]{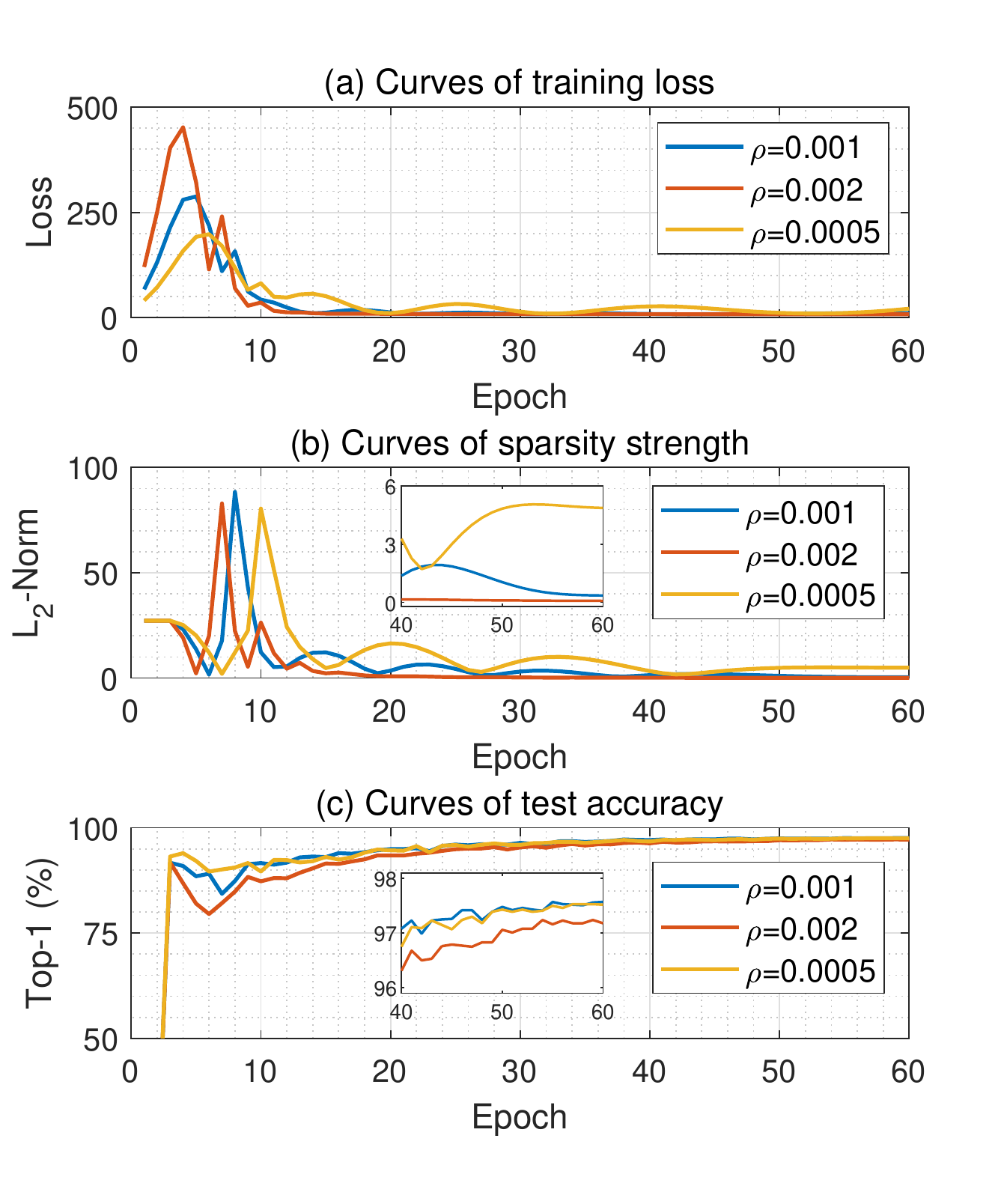}
    \vspace{-9mm}
    \caption{Effect of $\rho$ on the structured pruning procedure. $\rho$ controls the trade-off between the speed of imposing sparsity and task performance.
    %Too large $\rho$ leads to performance degradation.
    }
    \label{fig:ablation2}
\end{figure}

\textbf{Effect of Batch Size on Head Ranking.} As shown in Eq. \ref{eqn:cosine}, the importance scores of attention head is calculated on a batch of data. To investigate the potential impact of batch sizes for the ranking results, we observe the change of ranking with different batch sizes. As shown in Fig. \ref{fig:ablation3}, the ranking results are very stable (almost the same) when the batch size varies. Therefore we can conclude that using batches of data can already achieve very good estimation of head ranking. In other words, our ranking approach has low sensitivity to the distribution of input data. 

\textbf{Sensitivity of Penalty Parameter $\rho$.} We also explore the effect of hyperparameter $\rho$ on the structured pruning procedure. Fig. \ref{fig:ablation2} (a) shows the convergence of training process with respect to different $\rho$. It is seen that the convergence speed is always fast, and hence it demonstrates the promising convergence property of our approach in practice. Fig. \ref{fig:ablation2} (b) illustrates the $L_2$-norm of the masked entries. It is seen that the larger $\rho$ makes the model exhibit more sparsity at the earlier stage, thereby indicating that larger $\rho$ can bring fewer epochs in the final fine-tuning stage. However, as shown in Fig.~ \ref{fig:ablation2} (c), too large $\rho$ brings accuracy degradation, so $\rho$ can be considered as a parameter that controls the trade-off between the speed of imposing sparsity and task performance.

\begin{figure}[!h]
    \centering
    \includegraphics[width=0.99\linewidth]{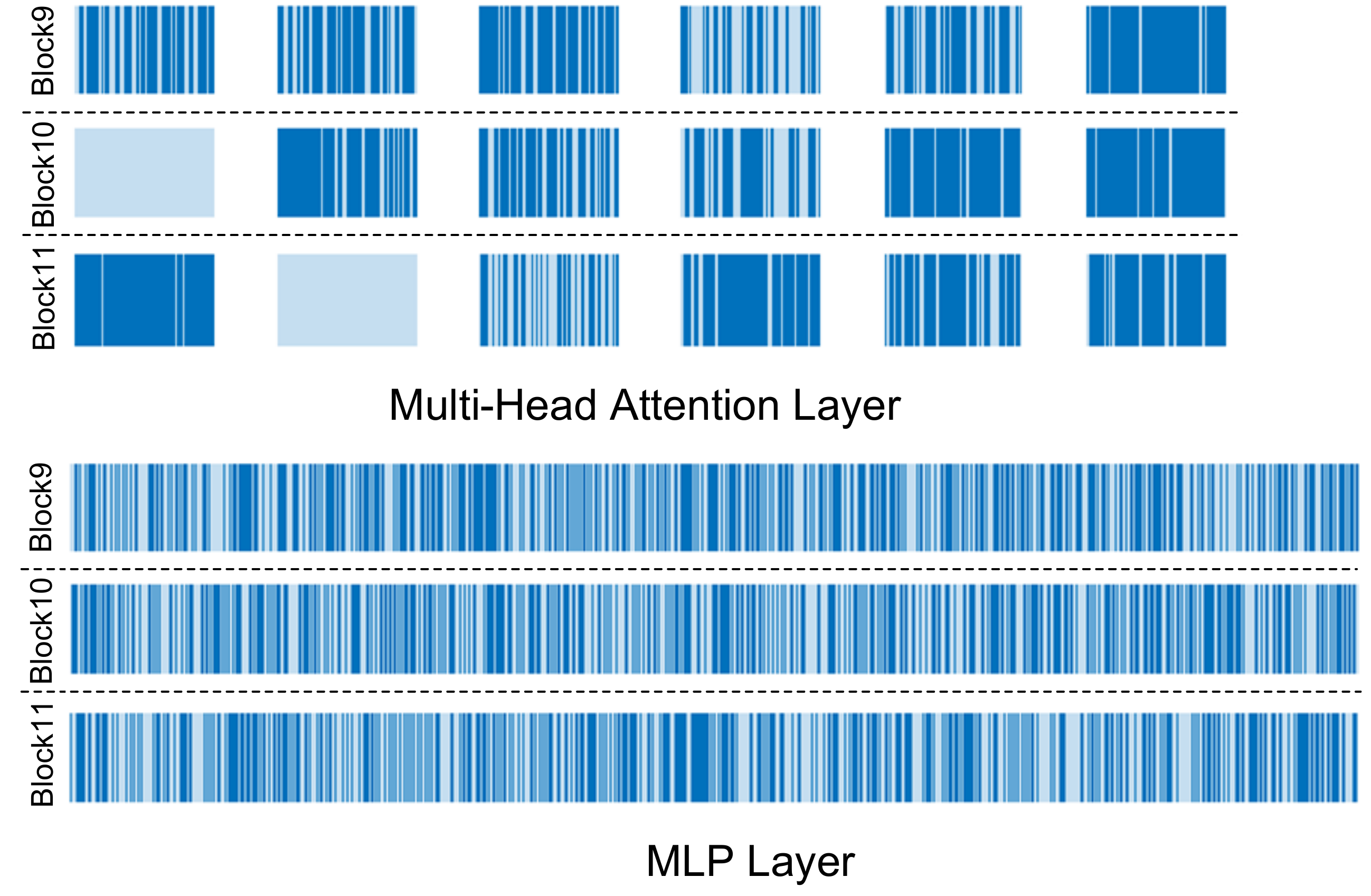}
    \vspace{-3mm}
    \caption{Visualization of the imposed structured sparsity on the DeiT-Small model. The columns and heads with lighter color are pruned. Our method can prune columns (Block9, Block10, and Block11), and heads (Block10, Block11) of the Multi-Head Attention layer. On the other hand, we can prune columns of MLP layers in all the blocks.}
    \label{fig:visual}
\end{figure}

\textbf{Visualization.} Fig.~\ref{fig:visual} illustrates the sparsity patterns in the pruned ViT models after performing our GOHSP approach. It is seen that three types of structured sparsity patterns (head-level sparsity, column-level sparsity in the head and column-level sparsity in the MLP) are imposed on the pruned models. Such pruning can be more effective on hardware than the unstructured pruning methods.

\textbf{Why Douglas—Rachford splitting method?} As shown in our Optimization section, the iterative Douglas—Rachford splitting technique is adopted to solve Eq. \ref{eqn:obj_2}. Such choice is built on two reasons. \underline{1) Convergence:} Douglas—Rachford splitting method is a primal-dual optimization method that enjoys fast convergence speed. According to \cite{boyd2011distributed}, within a few iterations it can provide satisfied solution  for large-scale problems -- particularly attractive for DNN applications. More specifically for this work, the fast convergence of Douglas—Rachford splitting method can avoid gradient explosion problem introduced by the additional sparsity loss in Eq. \ref{eqn:lagrangian}. \underline{2) Flexibility:} Douglas—Rachford splitting method, by its nature, divides the original difficult optimization problem into several less complicated sub-problems, each of which can be then addressed independently. This divide-and-conquer property is very suitable for optimizing the heterogeneous structured pruning of ViT, which explores the different types of structured sparsity across different attention heads and MLPs (Eq. \ref{eqn:update_w_attn} and \ref{eqn:update_w_mlp}).

% \underline{3) Performance:} Because of outstanding optimization performance, ADMM attracts increasingly attentions for optimizing DNN problems. For instance, ADMM-based pruning \cite{zhang2018systematic} outperforms APG-based pruning \cite{huang2018data} on the identical experiment.

\section{Conclusion}
In this paper we propose GOHSP, a unified framework to perform graph and optimization-based heterogeneous structured pruning for vision transformers.  By using graph-based ranking and leveraging the advanced optimization technique, our approach can efficiently impose different types of structured sparse patterns on the vision transformers with high compression rate and task performance. Our experiments show that, on ImageNet, with $30-50\%$ sparsity, GOHSP compresses the DeiT-Tiny and DeiT-Small models with minor or no loss in accuracy and with $\sim25$ improvement in rum-time efficiency. Finally, we compress ViT-Small up to $80\%$ on CIFAR10 with minor loss in accuracy.

\bibliography{aaai23}

\end{document}